%% file: main.tex
\pdfoutput=1
\documentclass[letterpaper, 10 pt, conference]{IEEEtran}  

\IEEEoverridecommandlockouts                              

\usepackage{graphics} 
\usepackage{epsfig} 
\usepackage{mathptmx} 
\usepackage{times} 
\usepackage{amsmath} 
\usepackage{amssymb}  

\usepackage[T1]{fontenc}
\usepackage{amsfonts}
\usepackage{booktabs}
\usepackage{siunitx}
\usepackage{caption} 
\usepackage{booktabs}

\usepackage{array}    

\usepackage{graphicx}
\usepackage{algorithm, algpseudocode}
\usepackage{float}
\usepackage[percent]{overpic}
\usepackage{caption}
\DeclareGraphicsExtensions{.jpeg,.png}

\usepackage[font=small,skip=0pt]{subcaption}

\usepackage{array, tabularx, makecell, booktabs}%

\usepackage{tikz}

\usepackage{geometry}
\usepackage{comment}

\usepackage[bookmarks=false]{hyperref}
\hypersetup{
    colorlinks=true,
    linktoc=all,
    citecolor=black,
    filecolor=black,
    linkcolor=blue,
    urlcolor=black
}

\usepackage{footmisc}

\usepackage{xcolor}

\title{\LARGE \bf
Arena 3.0: Advancing Social Navigation in Collaborative and Highly Dynamic Environments
}

\author{
    Linh K{\"a}stner$^{1,2,*}$, 
    Volodymyir Shcherbyna$^{1,*}$, 
    Huajian Zeng$^{3}$, 
    Tuan Anh Le$^{1}$, \\
    Maximilian Ho-Kyoung Schreff$^{1}$, 
    Halid Osmaev$^{1}$, 
    Nam Truong Tran$^{1}$, \\
    Diego Diaz$^{1}$, 
    Jan Golebiowski$^{1}$,
    Harold Soh$^{2}$, 
    and Jens Lambrecht$^{1}$ \\
    {\small $^{*}$Equal Contribution}\\
    {\small $^{1}$Technical University Berlin (TUB), Germany} \\
    {\small $^{2}$National University of Singapore (NUS), Singapore} \\
    {\small $^{3}$Technical University Munich (TUM), Germany}
}

\begin{document}

\maketitle
\thispagestyle{empty}
\pagestyle{empty}

\begingroup
\renewcommand\thefootnote{}
\footnotetext{\hspace*{-2.2em}This research / project is supported by A*STAR under its National Robotics Programme (NRP) (Award M23NBK0053). This research is supported in part by the National Research Foundation (NRF), Singapore and DSO National Laboratories under the AI Singapore Program (Award Number: AISG2-RP-2020-017).}
\endgroup

\input{0-abstract}
\input{1-introduction}
\input{2-Related-Works}

\input{3-methodology}

\input{4-evaluations}

\input{5-conclusion}
\input{acknowledgment}


\addtolength{\textheight}{-1cm} 




\typeout{}
\bibliographystyle{IEEEtran}
\bibliography{main}

\end{document}

%% file: 0-abstract.tex

\begin{abstract}
\noindent
Building upon our previous contributions, this paper introduces Arena 3.0, an extension of Arena-Bench \cite{kastner2022arena-bench}, Arena 1.0 \cite{kastner2021arena}, and Arena 2.0 \cite{kastner2023arena}. Arena 3.0 is a comprehensive software stack containing multiple modules and simulation environments focusing on the development, simulation, and benchmarking of social navigation approaches in collaborative environments. We significantly enhance the realism of human behavior simulation by incorporating a diverse array of new social force models and interaction patterns, encompassing both human-human and human-robot dynamics. The platform provides a comprehensive set of new task modes, designed for extensive benchmarking and testing and is capable of generating realistic and human-centric environments dynamically, catering to a broad spectrum of social navigation scenarios. In addition, the platform's functionalities have been abstracted across three widely used simulators, each tailored for specific training and testing purposes. The platform's efficacy has been validated through an extensive benchmark and user evaluations of the platform by a global community of researchers and students, which noted the substantial improvement compared to previous versions and expressed interests to utilize the platform for future research and development. Arena 3.0 is openly available at \href{https://github.com/Arena-Rosnav}{\color{blue}https://github.com/Arena-Rosnav}.

\end{abstract}

%% file: 1-introduction.tex
\section{Introduction}
\noindent
As the integration of human-robot collaboration becomes increasingly essential in fields such as healthcare, logistics, and delivery, the necessity for robots to navigate through dynamic and human-centric environments is paramount. The realm of social navigation, where robots maneuver and interact in human-populated settings, is rapidly gaining attention. Critical factors in navigation include not just safety, but also operational smoothness, user stress, acceptance, interaction, and maintaining efficiency. Recent years have seen strides in social robotics \cite{tsoi2022sean},\cite{francis2023principles},\cite{everett2018motion} with numerous research works proposing platforms and approaches for social navigation and benchmarking.
\begin{figure}[!h]
    \centering
    \includegraphics[width=0.99\linewidth]{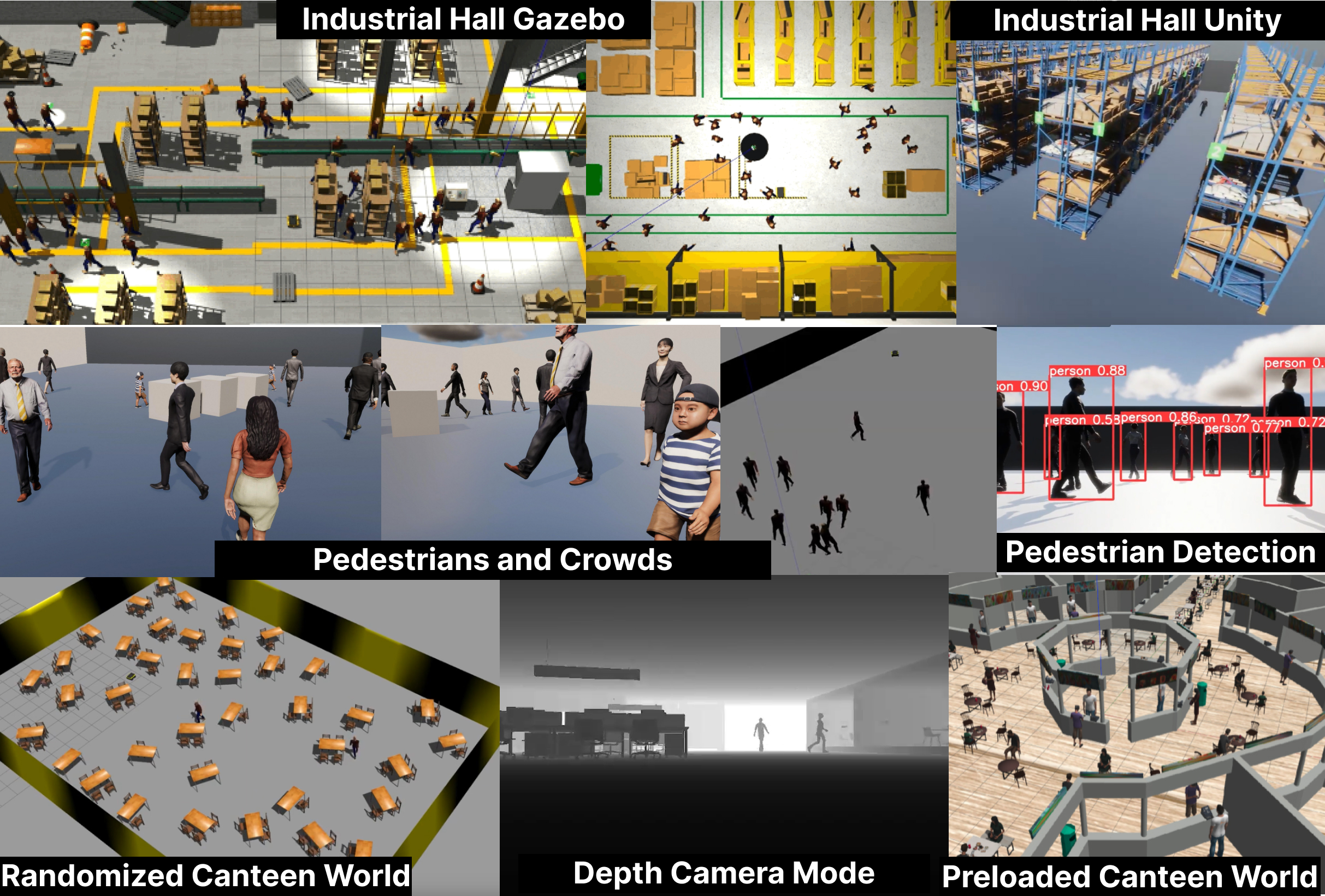}
    \caption{Sample scenes from the Arena 3.0 platform, which provides tools to develop social navigation approaches in highly dynamic and crowded environments. It focuses on social navigation and provides a number of modules to achieve realistic simulation of human-centric environments, developing and testing navigation algorithms on various robotic systems and setups, and simplified extension with new modules.}
    \label{intro}
\end{figure}
\\\noindent
However, many existing platforms are constrained to specific planning approaches, either learning-based or traditional, and exhibit limited extensibility \cite{francis2023principles}. Simulations often feature overly simplified human behavior models, leading to a widened simulation-to-reality (sim2real) gap. Furthermore, the focus on single simulators can be restrictive due to inherent limitations of individual simulators. Testing capabilities are also often constrained, with little environmental randomization or planner diversity.
\\\noindent
In our prior work Arena 1.0 \cite{kastner2021arena}, we tackled the challenge of deploying both learning-based and traditional planners within a unified simulation environment. We expanded this approach to include the deployment of 2D trained deep reinforcement learning (DRL) agents in Gazebo simulators, enhancing the realism of robotic kinematics and reducing the sim2real gap in Arena 2.0 \cite{kastner2023arena}. Additionally, we integrated an array of planners and an evaluation pipeline with Arena Bench \cite{kastner2022arena-bench}.
\\\\\noindent
However, in environments centered around human activity, robots must undergo rigorous testing for human-robot interaction. In these settings, realistic human simulation is crucial, which makes it necessary to not only model interactions among humans but also between humans and robots. These scenarios present new challenges and opportunities for robots to adapt their behavior and decision-making processes.
\\\\\noindent
With these considerations, we developed the third iteration of Arena to provide a comprehensive platform focusing on advancing social navigation. Arena 3.0 compromises an extensive software stack containing multiple modules and simulation environments. We incorporated state-of-the-art social force models and social interaction patterns. We have integrated and extended the Move Base Flex (MBF) navigation framework \cite{move_base_flex} and provided APIs for extending our modules. 
Additionally, we enhanced the task generator, which now includes tasks tailored for specific situations like blockages or emergencies, and introduce dynamically and randomly generated human-centric environments such as canteens, offices, or industrial halls, offering a vast array of training and testing environments. These functionalities are abstracted in the core module and are compatible on three simulators: Flatland, Gazebo, and Unity. \\\\
The main contributions of this work are the following:
\begin{itemize}
    \item Integration of human behavior models, enhancing modules with human and human-robot interaction forces.
    \item Development of an extended task generation toolkit, enabling users to create and design specific worlds, scenarios, and tasks for training and testing. This includes the dynamic and random generation of socially-centered environments such as canteens, warehouses, or offices.
    \item Abstraction of core functionalities across three widely used simulators: Flatland2D for efficient 2D training, Gazebo for realistic robot kinematics testing, and Unity for photorealistic training scenarios.
    \item Provision of comprehensive APIs, facilitating the straightforward extension of new modules, which includes social force models, social state machines, and intermediate planner modules.
    \item Enhancement of the robot and planning suite. The planning framework has been advanced to MBF \cite{move_base_flex}, which offers improved navigation performance and is extended by an intermediate planner concept for greater customization of user-developed planners.
\end{itemize}

\color{black}

%% file: 2-Related-Works.tex
\section{Related Works}
\noindent
This work builds on our previous platforms Arena-Bench \cite{kastner2022arena-bench}, Arena 1.0 \cite{kastner2021arena}, and Arena 2.0 \cite{kastner2023arena}, enhancing them for social navigation in dynamic, human-centric environments. Our initial focus was on learning-based obstacle avoidance and integrating classical with learning-based approaches. This latest version emphasizes environments for social navigation, a necessity due to the growing presence of robots in sectors like healthcare and public spaces.

\begin{figure*}[!h]
    \centering
    \includegraphics[width=0.99\linewidth]{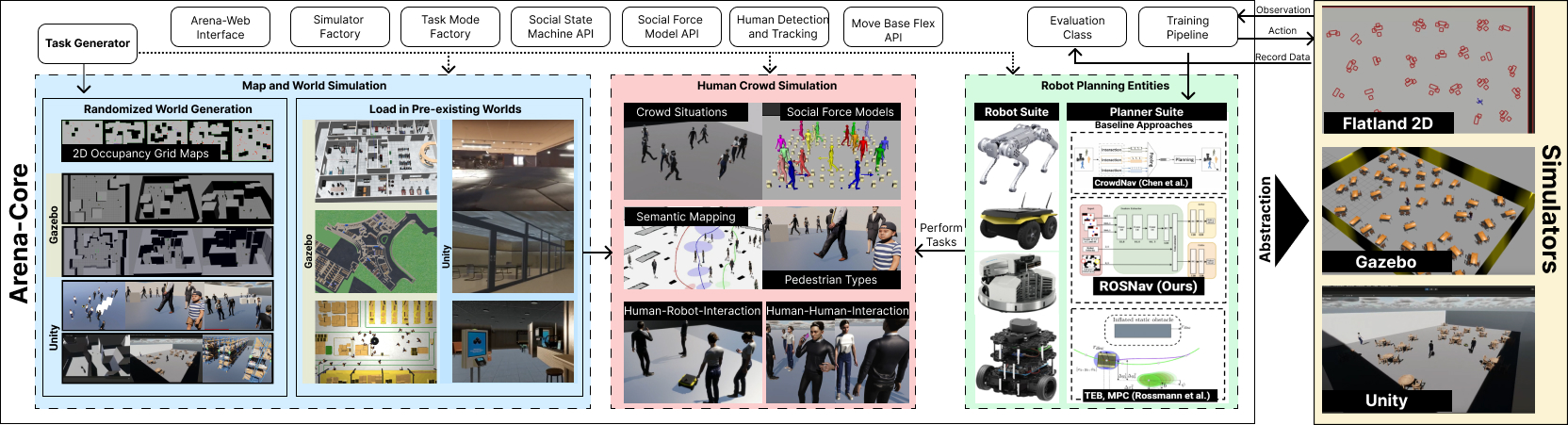}
    \caption{System architecture and modules of Arena 3.0. At its core, the platform consists of map and world generator algorithms, realistic pedestrian simulation, and comprehensive robot and navigation algorithm suites. The core architecture is fully abstracted from the simulators, allowing for cross-simulator scenario compatibility, with specific functionalities like RGB-D data available in Unity or Gazebo, and LiDAR applicable in all three simulators. Additionally, the platform features a training pipeline and an extended version of the MBF navigation framework to support the development and refinement of navigation approaches. Supplementary modules, such as the evaluation class and the web-based companion app, provide tools for data analysis and manual scenario creation. Using the provided API endpoints, the user can extend the platform with new planners (Move Base Flex API), task modes (task factory), social force models (SFM API), or simulators (simulator-factory). Further, computer vision modules for pedestrian detection and tracking are integrated to provide pedestrian data for further use.}
    \label{fig:system}
\end{figure*}
\noindent Social navigation is an emerging area of increasing importance in robotics and there have been a growing interest in developing methods \cite{everett2018motion}, \cite{dugas2020navrep}, \cite{francis2023principles}. For instance, works by Everett et al. \cite{everett2018motion}, Long et al. \cite{long2017deep} or Xiao et al. \cite{xiao2017human} implemented learning-based approaches for social norm adherence. Singamaneri et al. extended the classic TEB planner \cite{rosmann2015planning} with social metrics \cite{singamaneni2021human}. However, Francis et al.'s study \cite{francis2023principles} on principles and guidelines for social navigation research underscored the lack of comprehensive simulation platforms for navigation benchmarking.
\\\noindent
Platforms like SEAN 1.0 and 2.0 by Tsoi et al. \cite{tsoi2020sean, tsoi2022sean} provided valuable insights but were limited in terms of environment variety and simulation capabilities. Other contributions include NavRep by Dugas et al. \cite{dugas2020navrep}, Crowd-Robot Interaction by Chen et al. \cite{chen2019crowd} or Socially-Aware Navigation by Chen et al. \cite{chen2017socially}. Additional works in this realm include our previous works \cite{kastner2021arena, kastner-aio}, proposing various learning-based navigation strategies in crowded environments. These works however, focus mainly on developing learning-based navigation approaches, often in specialized simulation platforms and as such, were difficult to reproduce uniformly in other simulators or experimental setups.
\\\noindent
Benchmarks like Bench-MR by Heiden et al. \cite{bench_mr}, Robobench by Weisz et al. \cite{robobench}, CommonRoad by Althoff et al. \cite{althoff2017commonroad}, and the Benchmarking suite by Moll et al. \cite{moll2015benchmarking} contributed significantly to static environment navigation but did not address dynamic human-robot interaction complexities. Tsoi et al.'s SEAN platform \cite{tsoi2020sean, tsoi2022sean}, SocialGym by Holtz et al. \cite{holtz2021socialgym}, HuNavSim by Perez-Higueras et al. \cite{perezhunavsim}, and MRPB 1.0 by Wen et al. \cite{mrpb} made strides in social navigation but faced limitations in scenarios or robot variety.
\\\noindent
Furthermore, works like Interactive Navigation by Xia et al. \cite{xia2020interactive}, ROS Navigation in Cluttered Environments by Portugal et al. \cite{portugal2019ros}, and the Social Evaluation Platform by Gao et al. \cite{gao2022evaluation} have contributed to the understanding of navigation in complex dynamic environments, however are also limited in functionality and selection of robots and planners. The BARN challenge by Xiao et al. \cite{xiao2022autonomous} and DynaBARN by Nair et al. \cite{nair2022dynabarn} also aim to p, though with certain limitations in environment diversity and a have simplified obstacle simulation.
Our proposed platform addresses aforementioned gaps and aspire to provide a platform with realistic human and crowd behavior, comprehensive suites of robots and planning approaches, and extensive toolkits for highly diverse scene and world generation.

%% file: 3-methodology.tex
\section{Overview of Arena 3.0}
\noindent Arena 3.0 consists of several key modules, which include pedestrian simulation, world and task generation, a training pipeline, and the integration of planner and robot suites. It features specific API endpoints for the integration of self-developed modules and supplementary tools like the evaluation class or a web-based front end for manual scenario creation. 

\subsection{System Design and Differences to the Previous Versions}
\begin{figure*}[!h]
\centering
\includegraphics[width=0.99\linewidth]{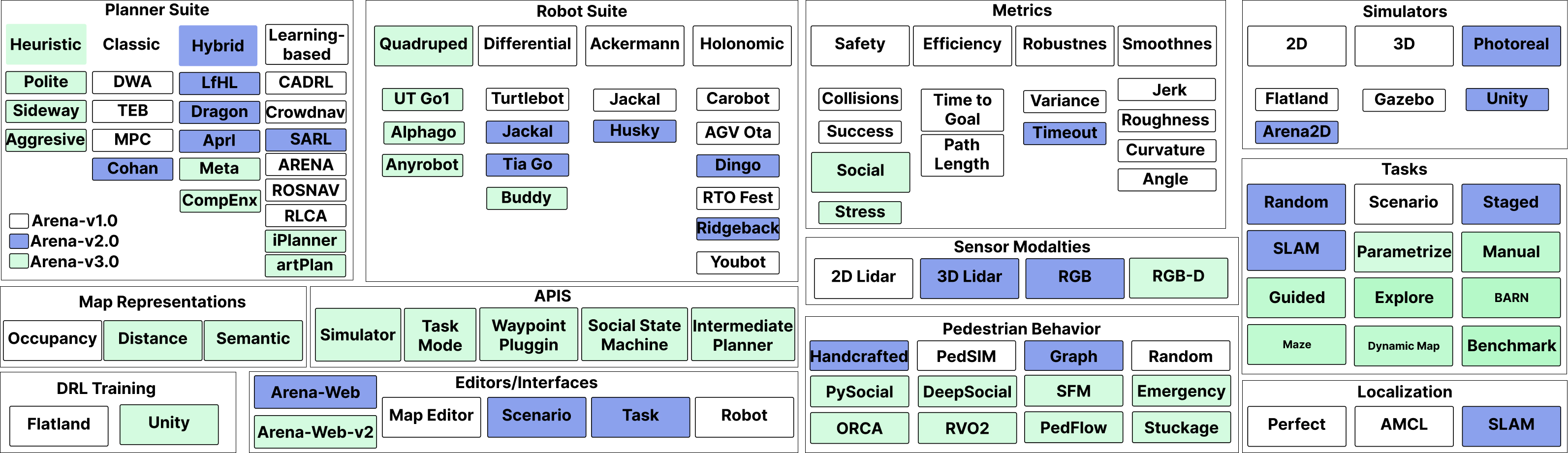}
\caption{Differences between Arena versions: Arena 1.0 \cite{kastner2021arena} (white), Arena 2.0 \cite{kastner2023arena} (blue), and Arena 3.0 (green). Key enhancements include the complete abstraction of the Arena core, substantially extended map and world generation algorithms and task modes, additional robot and planners, provision of APIs for extensions, and a focus on social navigation, notably multiple realistic crowd behavior models, human-human and human-robot interactions, and social metrics.}
\label{fig:diff}
\end{figure*}

\noindent Figure \ref{fig:system} outlines the current version's modules, while Figure \ref{fig:diff} provides an overview of differences with previous versions of the platform. Arena 3.0 introduces a core module, denoted as Arena-core, that abstracts functionalities from three simulators, enabling most functions to run concurrently across these simulators (subject to limitations like RGB-D being exclusive to Unity and Gazebo).
While all simulators were already available and could be selected in the previous version of Arena, only basic functionalities such as loading an empty world, spawning a robot, or loading obstacles were provided and the user had to implement further functionalities for each simulator individually. In contrast, Arena 3.0 introduces the complete abstraction of all functions to completely automate processes making it possible to include new functions across all simulators simultaneously by only extending the core module.
This design also enhances comparability of approaches and interoperability, e.g., for algorithms trained in a 2D simulator and later validated in Gazebo. A vital component of this system is the crowd and human simulation, incorporating various advanced social force models, social interactions patterns, and varying human types.

\noindent Using the improved map and task generators, users can generate diverse environments in socially-centric settings like canteens, offices, or warehouses, with each episode offering variation through our map generators. Pre-built worlds, such as a replica of the National University of Singapore campus or hospitals, are also available for loading. This variety enables users to design highly specific or randomized scenarios for both qualitative and quantitative testing and validation. Furthermore, the robot and planner suite was extended offering a wide range of options. Additional modules, like the extended navigation stack and training pipeline, are provided to develop and fine-tune navigation strategies. Furthermore, we also included computer vision approaches such as YOLO \cite{redmon2016you} to detect and track humans. Additionally, the companion Webapp Arena-web \cite{kastner2023arenaweb} has been extended and adapted significantly to ensure that all new functions and modules of Arena 3.0 were compatible with the Webapp. It is thus denoted as Arena-Web-v2.

\subsection{Human Crowd Behavior Modeling and Simulation}
\noindent
A primary objective of Arena 3.0 is the enhancement of crowd and human simulation realism. While a range of social force models exists, many simulators still rely on simpler models like ORCA or RVO2, or the older Social Force Model (SFM) of Helbing et al. \cite{helbing1995social}. Recent years have seen the development of newer models, but these efforts have largely been isolated to individual research groups and have not gained widespread adoption. In response, Arena 3.0 aims to integrate a comprehensive range of these models. Table \ref{tab:sfm} lists all the SFMs incorporated into Arena 3.0.
\begin{table}
    \centering
    \scriptsize 
    \caption{Overview of integrated social force models.}
    \label{tab:sfm}
    \setlength{\tabcolsep}{2pt} 
    \renewcommand{\arraystretch}{1.1} 
    \begin{tabular}{p{2cm}p{6.3cm}}
    \hline 
    \textbf{Plugin} & \textbf{Description} \\
    \hline
    \hline
    Pedsim \cite{helbing1995social} & Base social force model of Helbing et al. \\
    \hline
    Passthrough & Unaltered Pedsim\_ros behavior using the SFM of Helbing et al. \\
    \hline
    Spinny & Pedestrians moving in a circle. Used for demonstration purposes. \\
    \hline 
    PySocial \cite{pysocialforce} & Models group behavior with a modular configuration. Modular and extendable, natural movement in less dense situations. \\
    \hline 
    Deep Social \cite{deepsocialforce} & Trained with video data of real crowds. However, slow movement and reaction times of pedestrians to fast incoming robots. \\
    \hline 
    Evacuation \cite{sfm-evacuation} & Simulates evacuation scenarios with a strong repelling force among pedestrians and an uniform goal for all. \\
    \hline 
    Bonding \cite{sfm-bonding} & A variation of the evacuation model including a bonding term. Pros: High number of adjustable parameters. Cons: Requires careful adjustment. \\
    \hline
    ORCA \cite{van2011reciprocal} & No realistic movements but rather obstacle avoidance for simple use cases. \\
    \hline
    RVO2 \cite{van2011rvo2} & RVO2 extension of ORCA for obstacle avoidance. \\
    \hline
    \end{tabular}
\end{table}
\\\noindent
Given its ease of extension, robustness, and the breadth of functionalities, we adopted the PySocialForce model as the default force model in Arena 3.0, with other models built upon its framework. PySocialForce is an implementation of the extended SFM by Moussaid et al. \cite{sfm-moussaid}, developed by Gao et al. \cite{pysocialforce} and made available as open-source software. In previous versions of Arena, we utilized a simpler social force model based on Helbing et al.'s work \cite{sfm-helbing}, implemented in Pedsim. Using PySocialForce and other integrated models marks a significant step in advancing the fidelity and applicability of human behavior simulation in robotic navigation contexts.
\\\noindent
The previous SFM models the behavior of pedestrians through multiple forces, originating from different sources, acting on the agent. These sources include the goal attraction $\vec{f}^0$, representing the agent's desire to reach a specified waypoint, the pedestrian repulsion $\vec{f}_{ij}^p$ between agent $i$ and agent $j$, modeling the behavior of an agent towards other agents, and the border repulsion $\vec{f}^{b}$, modeling the agent's repulsion from buildings or obstacles. When calculating the absolute force acting on an agent $\vec{f}_i$, the different forces are aggregated via a sum of all forces: \\
\begin{equation} \label{eq:absolut_force}
    \vec{f}_i = \vec{f}^0_i + \sum_{j}{\vec{f}_{ij}^p} + \vec{f}^{b}_i
\end{equation}
This describes the equation of motion since $f_i$ will be equal to the acceleration:
\begin{equation}
    \frac{d \vec{v}_i}{dt} = \vec{f}_i
\end{equation}
\\\noindent
The extended social force model by Moussaid et al. \cite{sfm-moussaid} additionally models the behavior of groups, adding 3 additional forces. These forces model how members of the same group interact with each other. The model includes a group gaze force $\vec{f}^{gaze}$, that models a group members behavior to look in a direction where it can see other group members. Also, the model includes an attractive force $\vec{f}^{attr}$ that acts towards the center of mass of the group. Correspondingly, there is also a repulsive force $\vec{f}^{rep}$, stopping group members from colliding. They extend equation \ref{eq:absolut_force} to:
\begin{equation}
    \vec{f}_i = \vec{f}^0_i + \sum_{j}{\vec{f}_{ij}^p} + \vec{f}^{b}_i + \vec{f}^{gaze}_i + \vec{f}^{attr}_i + \vec{f}^{rep}_i
\end{equation}
\\\noindent
Although this extended social force model enables the modeling of groups in our simulation, we saw a slight decrease in object avoidance ability, e.g. when the paths of different groups overlap.  \\
To mitigate this issue, we made a number of adjustments to the original model. In order to not disrupt the group dynamic, we overwrite the waypoint of each pedestrian to match the waypoint of a randomly assigned `group leader'. 
\begin{figure*}[!h]
    \centering
    \includegraphics[width=0.9\linewidth]{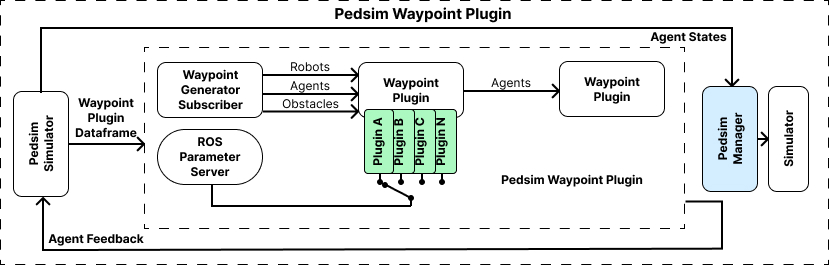}
    \caption{System design of the Pedsim Waypoint Plugin. Using this API enables users to seamlessly integrate new social force models and pedestrian behaviors, facilitating the inclusion of complex calculations and behaviors.}
    \label{fig:wp}
\end{figure*}
\\\noindent
Furthermore, the Pysocial Force implementation, uses a mechanism where the force of an agent, upon reaching the waypoint or being within a close distance, inverts the force with every update. This makes the agent spinning around its waypoint. We evade this problem by immediately updating the waypoint to the next one with a more tolerant threshold when such a waypoint is 'reached'. As a result, our optimized Pysocial Force implementation produces realistic pedestrian and crowd simulations, which are visually presented in the supplementary video.

\subsection{The Pedsim Waypoint Plugin}
\noindent
The simulation of crowd behavior is driven by SFMs, with movement patterns communicated via ROS topics. The coordination, visualization, and synchronization of crowd actions within the simulation environment are managed using the Pedsim ROS package. This base package has been extended by our \emph{Pedsim Waypoint Plugin} interface, illustrated in \autoref{fig:wp}, which is a waypoint-centric approach to control the navigation behavior of pedestrians and pedestrian groups within the simulation. To facilitate the integration of new SFMs or specific behaviors into the base model, we have developed a Pedsim Waypoint Plugin API. This API allows users to easily incorporate their custom social force models or behaviors, enhancing the versatility and applicability of the simulation. 
\\\noindent
Overriding Pedsim calculations is achieved by forwarding the output of the Pedsim simulator to an additional Pedsim agents node, which acts as part of a feedback loop and sends physical forces for the actors back to the Pedsim simulator node. This is realized by an event-driven, control-loop architecture,  which is abstract by design to allow for various approaches to computational scalability challenges, such as stateful computations and output interpolation. In order to allow complex calculations e.g. as required by some other SFMs, the basic simulated agents data is complemented by additional pre-calculation data and packaged into a Pedsim Agents Dataframe message.
The pre and post-processing of the full messages coming in and out of the plugin node is handled automatically, and the raw data is handled inside a pre-selected plugin. This approach not only preserves the original functionality of Pedsim but extends it, allowing for more complex and nuanced simulations of crowd behavior and dynamics.

\subsection{Social States and Semantic Mapping}
\noindent In addition to the SFMs, which model the movement and behavior of pedestrians in crowds, we also extend the plugin with social states to contain particular human-human-interactions (HHI) and human-robot-interactions (HRI). Table \ref{tab:states} lists the currently available social states of pedestrians. 
\begin{table}
    \renewcommand{\arraystretch}{1.5}
    \scriptsize 
    \caption{An overview of the implemented social states and their respective animations. HOI: Human Object Interaction, HHI: Human-Human-Interaction, HRI: Human-Robot-Interaction}
    \label{tab:states}
    \centering
    \begin{tabular}{m{2,5cm}m{0,5cm}m{4,2cm}}
    \hline 
    Social States & Type & Description \\
    \hline\hline 
         Standing & HOI & Person standing idle, will move once objects approaches \\ \hline 
         Walking/ Running & HHI & Person running, velocity range set in config \\ \hline 
         Robot Avoidance & HRI  & Humans avoid the robot slowly \\ \hline
         (Group-) Talking/ Listening/ Telling a Story & HHI & multiple pedestrians gather and talk \\ \hline 
         Texting & HOI & Humans looking at the phone and text\\ \hline 
         Interested, obstacle interaction & HRI & Humans approach robot and playing with it \\ \hline
         Talking on the phone & HOI & Human talsk on the phone \\ \hline
    \end{tabular}
\end{table}
\noindent
With these social states and various forms of interactions, additional information such as semantic information can be provided for further use such as to enhance the observation space of an agent. Therefore, we introduce the concept of semantic layers, storing semantic data in a structured manner. This is achieved through the implementation of a Semantic Layer Plugin, which complements the ROS costmap package.\\
The Semantic Layer Plugin offers a general framework for storing and utilizing semantic data in conjunction with the costmap. The generation of semantic data originates from our Pedsim ROS implementation. Consequently, the encoding of this semantic data is centered around the Pedsim ROS container, characterized by a simple structure:

\begin{itemize}
    \item geometry\_msgs/Point location,
    \item float32 evidence.
\end{itemize}

\noindent This abstract design allows the evidence value to encode various semantics, essentially acting as a versatile numerical representation. Multiple instances of Pedsim ROS are compiled into a timestamped and string-labeled semantic data aggregate. The semantic data, encapsulated in PedsimData topics, can be directly given as input into the semantic layers of a costmap. The costmap, in turn, stores and maintains the latest PedsimData messages, tracking their currency and availability. In practical applications, this semantic data is utilized for both discrete information (such as type and state) and continuous data (velocity and observed probability).

\section{Robot Navigation Suite}
\noindent
Arena 3.0 encompasses a diverse array of robot platforms covering the main robot kinematics such as holonomic, differential drive, car-like, or quadruped robots.
Table \ref{tab:suite} provides an overview of the available robots and planners, along with their compatibility and descriptions.

\begin{table}[!h]
    \renewcommand{\arraystretch}{1.5}
    \scriptsize 
    \centering
    \caption{Overview of the robot and planner suites with descriptions of the planners and indications on which robot platform they can be deployed. }
    \begin{tabular}{m{1.0cm}m{0.8cm}m{1.0cm}m{4.2cm}}
    \hline 
    Planner & Type & Robot & Description \\
    \hline\hline 
    Applr \cite{xiao2020appld} & Hybrid & Jackal & A hybrid planner combining different approaches for adaptive planning. \\
    \hline 
    Cohan \cite{singamaneni2021human} & Classic & All & A traditional planner focusing on human-aware navigation strategies. \\
    \hline 
    Dragon \cite{xiao2022autonomous} & Hybrid & Jackal & A hybrid navigation system designed for dynamic environments. \\
    \hline
    DWA \cite{khatib1986real} & Classic & All & Dynamic Window Approach, a reactive collision avoidance method considering the robot's dynamics. \\
    \hline 
    TEB \cite{rosmann2015timed} & Classic & All & Timed Elastic Bands, optimizing a global path by considering kinematic and dynamic constraints. \\
    \hline 
    MPC \cite{rosmann2019time} & Classic & All & Model Predictive Control, using a model of the robot's dynamics to predict and optimize future trajectories. \\
    \hline
    LfLH \cite{xiao2022motion} & Hybrid & All & A hybrid planner leveraging both learned and heuristic components for effective navigation. \\
    \hline
    RLCA \cite{long2018towards} & Learning & All & Reinforcement Learning Collision Avoidance, using RL for dynamic obstacle avoidance. \\
    \hline
    ROSNavRL (OURS) & Learning & All & A learning-based approach trained on Arena 2.0 with Reinforcement Learning. \\
    \hline
    TRAIL \cite{xiao2022autonomous} & Learning & All & A learning-based planner focusing on trail navigation in unstructured environments. \\
    \hline
    Crowdnav \cite{chen2019crowd} & Learning & All & Focuses on navigating safely and efficiently in crowded environments using machine learning techniques. \\
    \hline
    SARL \cite{li2019sarl} & Learning & All & Socially Aware Reinforcement Learning, emphasizing social norms in navigation. \\
    \hline
    BRNE \cite{muchen23} & Game theory & Quadruped & Utilizes game theory for decision-making in complex scenarios, suitable for quadruped robots. \\
    \hline
    iPlanner \cite{yangiplanner23} & Learning & Quadruped & A learning-based planner for complex dynamic situations, suitable for quadruped robots. \\
    \hline
    ArtPlanner \cite{wellhausen2021rough} & Learning & Quadruped & A learning-based planner optimized for articulated, quadruped robots. \\
    \hline
    Aggressive (OURS) & Heuristic & All & Employs aggressive strategies for rapid navigation, often in competitive or urgent scenarios. \\
    \hline
    Polite (OURS) & Heuristic & All & Prioritizes polite and socially compliant behaviors in navigation. \\
    \hline
    Sideways (OURS) & Heuristic & All & Specializes in sideways maneuvers, useful in narrow or constrained spaces. \\
    \hline
    \end{tabular}
    
    \label{tab:suite}
\end{table}

\noindent We have upgraded the navigation framework by adopting MBF \cite{move_base_flex}, which is an enhancement of Move Base \cite{move_base}, effectively backporting features from the ROS 2 \cite{ros2} navigation stack to ROS. This forms the modern foundation of our navigation system, to which we have added a modified version of the intermediate planner (interplanner) concept from our previous works \cite{kastner2021connecting}. The interplanner serves as a distinct stage in the data flow process, bridging the global planner and the local controller. It receives a global path from the planner and publishes a refined intermediate path to the controller. The key advantage of the interplanner lies in its dual access to both local and global costmaps, enabling it to effectively subsample the global path based on local data, while also modifying controller parameters from a broader perspective. To showcase the capabilities of our API, we developed three heuristic planners, each illustrating specific robot behaviors in crowds: aggressive, polite, and sideways, using our intermediate plugin. These behaviors activate when the robot's Euclidean distance to dynamic obstacles falls below a defined threshold, prompting the intermediate planners to alter the robot's behavior accordingly. The planner's behaviors are visually demonstrated in the accompanying supplementary video.

\section{The Task Generator}
\noindent
The task generator is an essential component of the Arena core and was enhanced and restructured substantially. It is responsible for a broad variety of functionalities, including loading scenarios, generating static and dynamic obstacles, updating robot goals, interfacing with RVIZ, and initiating robot controllers and model loading into the simulator. 
\\\noindent
It also introduces independent task modes for both robots and obstacles, contributing to a more flexible and dynamic simulation environment. Table \ref{tab:modes} lists and describes all the currently available task modes. One of the main task modes is the random mode, which will spawn obstacles randomly given a predefined range. Another important mode is the scenario mode, which caters more to qualitative evaluations by providing consistent settings that allow for the examination of planners under uniform conditions. A new task mode that was added is the benchmark task mode, which is particularly valuable for facilitating unified competition and benchmarking. In this task, users can specify a single configuration file that outlines a series of task modes and the number of episodes for each. This configuration file can then be used by other users to execute planner on the exact same task mode configurations. To add new user specific task modes, the user is provided the task mode factory class API.
\begin{table}
    \caption{Overview of all task modes.}
    \centering
    \scriptsize 
    \setlength{\tabcolsep}{2pt} 
    \renewcommand{\arraystretch}{1.1} 
    \begin{tabular}{m{1.5cm}m{1.5cm}m{5cm}}
        \hline
        \textbf{Parameter} & \textbf{Task modes} & \textbf{Explanation} \\ 
        \hline\hline
        Obstacle (tm\_obstacles) & Scenario & Loads obstacles described in scenario JSON file. \\ 
                                 & Random & Loads random obstacles defined in task\_manager.yaml file. \\
                                 & Parametrized & Loads random obstacles for each model specified in XML file. \\
        \hline
        Robots (tm\_robots) & Scenario & Set start and goal for robot described in scenario JSON file. \\
                            & Random & Set random start and goal for robot. \\
                            & Guided & Changes RViz tool (2D Nav Goal) to set cyclic waypoints for robots and (2D pose estimate: reset robot position and sequence of waypoints). \\
                            & Explore & Obstacles are never reset and the robot receives random goals. \\
        \hline
        Modules (tm\_modules) & Staged & Allows to switch between stages defined in training curriculum file. \\
                              & Dynamic & Creates a dynamic map. \\
                              & RViz UI & Overwrites RViz Tools (2D Nav Goal: change goal of robot, 2D Pose Estimate: change robot current position, Publish Point: manual reset). \\
                              & clear forbidden zones & Clears forbidden zones for spawning obstacles. \\ 
                               & Benchmark & The user can provide a config to stage different task modes together. The benchmark config acts as a challnge or benchmark that other user can try and assess their planners. \\ 
        \hline
    \end{tabular}
    \\
    \label{tab:modes}
\end{table}

\begin{figure*}[!h]
    \centering
    \includegraphics[width=0.99\linewidth]{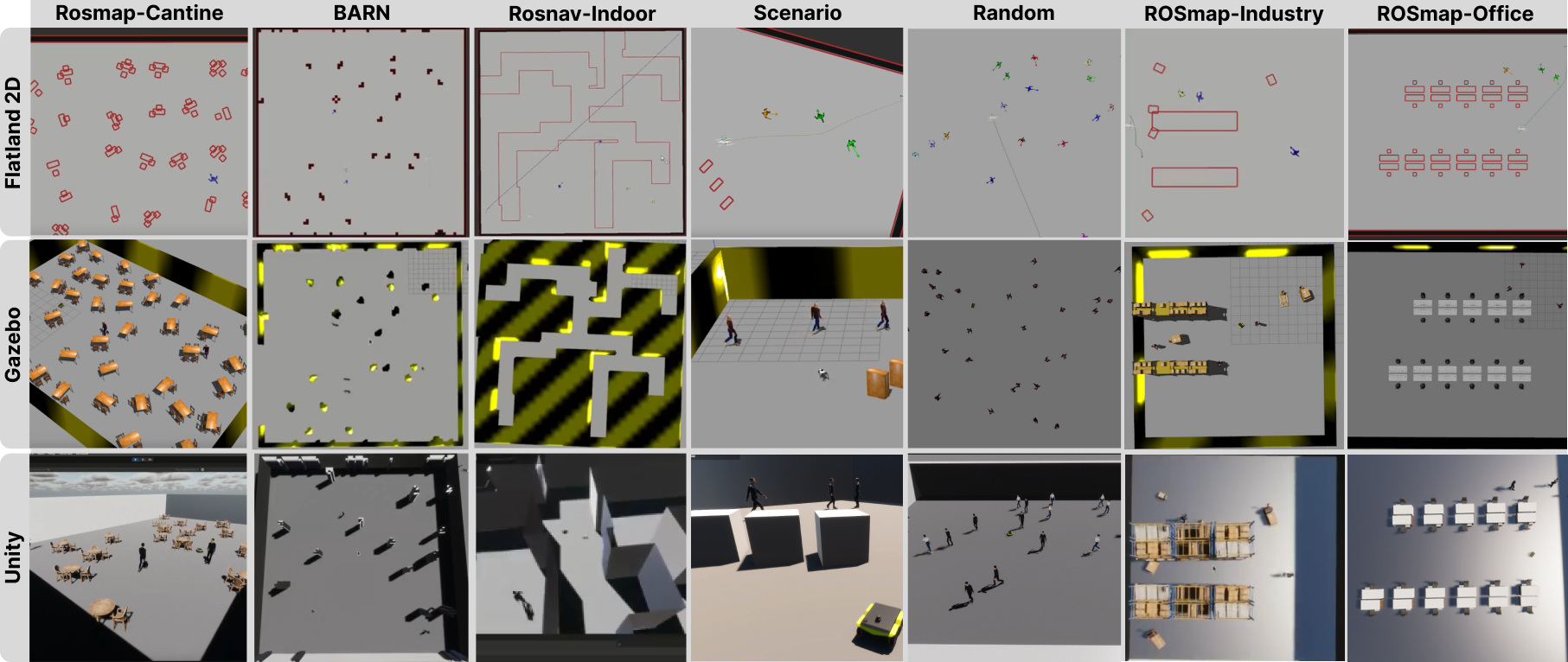}
    \caption{Different map and world generator modes on three simulators. The map and world generation is abstracted from the simulators making the creation of worlds and scenarios unified across all simulators. This provides the ability to train and test the same agents on different simulators leveraging the strength of each simulator.}
    \label{fig:tasks}
\end{figure*}

\section{World and Map generation}
\noindent
Closely integrated with the Task Generator in Arena 3.0, are the map and world generators, each serving a unique function in the creation of simulation environments. The map generator is responsible for producing a 2D occupancy grid and a corresponding distance map, essential for navigation and planning in two-dimensional space. Conversely, the world generator focuses on the requirements of 3D simulations, transforming the 2D maps into immersive 3D worlds.
\\\noindent
We have implemented a variety of map generation algorithms that are capable of generating dynamic and randomized worlds offering an extensive variety of environments for both training and testing. Users can generate different kinds of map designs such as indoor or outdoor maps, or more specialized scenarios like canteens, offices, or warehouses while still maintaining domain randomization. This capability allows for the creation of countless variations within a particular environment type. For instance, within the category of canteens, the system can generate thousands of distinct worlds, each with slight variations. This feature is crucial for providing unlimited numbers of environments and scenarios for testing and training purposes. \\\noindent
The latest version of our platform includes seven different map/world generation algorithms:
\begin{itemize}
    \item \textbf{Barn:} Generates a rectangular map outlined by walls and being empty inside. Afterwards the map is filled by small groups of walls that serve as obstacles. (Configurable Params: fill\_pct [how much of the map should be filled], smooth\_iter [roundness of wall groups])
    \item \textbf{Rosmap Outdoor:} Generate a rectangular map outlined by walls. Place trees at random positions without overlapping. (Configurable Params: obstacle\_num [number of tables to be spawned], obstacle\_extra\_radius [block a square space with edge length $1+2\cdot obstacle\_extra\_radius$ with the tree in the center])
    \item \textbf{Rosmap Indoor:} Generate a structure with greater complexity. This is done by generating rectangular rooms and connecting them by corridors to the nearest room. The rooms can overlap. (Configurable Params: corridor\_radius [determines the width of the corridor by $1+2\cdot corridor\_radius$], iterations [number of rectangular rooms to be generated])
    \item \textbf{Rosnav Outoor:} Generate a rectangular map outlined by walls. Then block certain square spaces using walls. These spaces can overlap with each other. (Configurable Params: obstacle\_num [number of blocking spaces to place], obstacle\_extra\_radius [determines the size of the blocking space by $1+2\cdot obstacle\_extra\_radius$])
    \item \textbf{Rosnav Canteen:} Places a table at a random free position with enough space to other chairs. This table can have up to 4 chairs surrounding it, but each chair has only a certain probability to be spawned, i.e. if the probability is 0.5 then the probability that a table has 4 chairs is 0.0625. (Configurable Params: obstacle\_num [number of tables to be spawned], obstacle\_extra\_radius [block a square space with edge length $1+2\cdot obstacle\_extra\_radius$ with the table in the center], chair\_chance [Chance for each chair at a table to spawn])
    \item \textbf{Rosnav Industrial Hall:} Places a shelfs at a random free position with enough space to other shelfs. The configuration of the shelfs follows a horizontal or vertical line. (Configurable Params: obstacle\_num [number of shelfs to be spawned], rack\_chance [Chance for each rack near a shelf to spawn])
    \item \textbf{Rosnav Office:} Places a workplace consisting of a table, a chair and a seperator at a random free position with enough space to workspaces. The configuration of the shelfs follows a horizontal or vertical line. (Configurable Params: obstacle\_num [number of workplaces to be spawned])
    
\end{itemize}

\subsection{Calculation of Evaluation Metrics}
\noindent
One key objective of Arena 3.0 is the benchmarking and evaluation of planners. As such, the platform includes a comprehensive data recording feature. This data recorder can be activated via a command line argument and systematically captures essential data for analysis.
\\\noindent
As the emphasis of Arena 3.0 is on social navigation, we also provide three new metrics to assess the social performance of each planner and robot. 
Studies by Rubagotti et al. \cite{rubagotti2022perceived} highlight the psychological impact of robots on humans, noting that prolonged direct facing by robots can elevate stress and anxiety levels among individuals. In response, we measure the duration a robot facing a pedestrian and vice versa, providing insight into the planner's effectiveness in maintaining comfortable interactions.
\\\noindent
Additionally, research by Howell et al. \cite{howell2023effects} and Wang et al. \cite{wang2019learning} indicates that robots operating within close proximity to humans can trigger feelings of uncanniness and reduce user acceptance. To address this, we utilize a 'private zone' defined by a 0.5m radius around humans and record the robot's time spent within this zone, serving as an indicator of the planner's ability to respect personal space and enhance user comfort.
Table \ref{tab:metrics} lists relevant metrics provided by Arena 3.0. 

\begin{table}
\centering
	\setlength{\tabcolsep}{0.2pt}
	\renewcommand{\arraystretch}{0.5}
		\caption{Overview of evaluation metrics}
	\begin{tabular}{lcp{3.5cm}}
		\hline
		Metric  &Unit & Explanation    \\ \hline
		Success Rate$^{2}$ & \%        & Runs with < 2 collisions          \\ 
		Collision$^{1,2}$ & -        & Total number of collisions\\
		Time to reach goal$^{2}$& [$s$] & Time required to reach the goal   \\ 
		Path Length$^{1,2}$  & [$m$]    & Path length in m           \\ 
		Velocity (avg.)$^{2}$     & [$\frac{m}{s}$]  & Velocity of the robot \\
		Acceleration (avg.)$^{2}$ & [$\frac{m}{s^2}$] & Acceleration of the robot \\
		Movement Jerk$^{2}$ & [$\frac{m}{s^3}$] & Derivation of Acceleration \\ 
		Curvature(avg.,max.,min.,norm.)$^{2}$ & [$m$] &  Degree of trajectory changes\\
		Angle over length$^{2}$  & $[\frac{rad}{m}]$ & Curvature over the path-length \\
		Roughness$^{2}$ & -                 & Quantifies trajectory smoothness \\ 
		Time in private zone$^{1}$ & $[s]$ & Time the robot spent in private zone \\
  		Time facing pedestrians$^{1}$ & $[s]$ & Time robot direction faced peds\\
        Time seen by pedestrians$^{1}$ & $[s]$ & Time peds directions faced robot's\\
		\hline
	\end{tabular}
    \footnotesize{$^1$Quantitative metric, $^2$Qualitative metric}\\
	\label{tab:metrics}
	
\end{table}

%% file: 4-evaluations.tex
\section{Validation and Evaluation}
\noindent
To evaluate the usability and functionality of our platform as well as capture general user opinions and the potential for further advanced usage of the platform, we conducted a study asking participants to install the platform, test out specific modules, and fill out a questionnaire. To demonstrate the benchmarking capabilities of using the Arena platform, we conducted an extensive evaluation run with a variety of planners using the platform and recorded data for evaluation. 

\begin{figure*}[!h]
    \centering
    \includegraphics[width=0.99\linewidth]{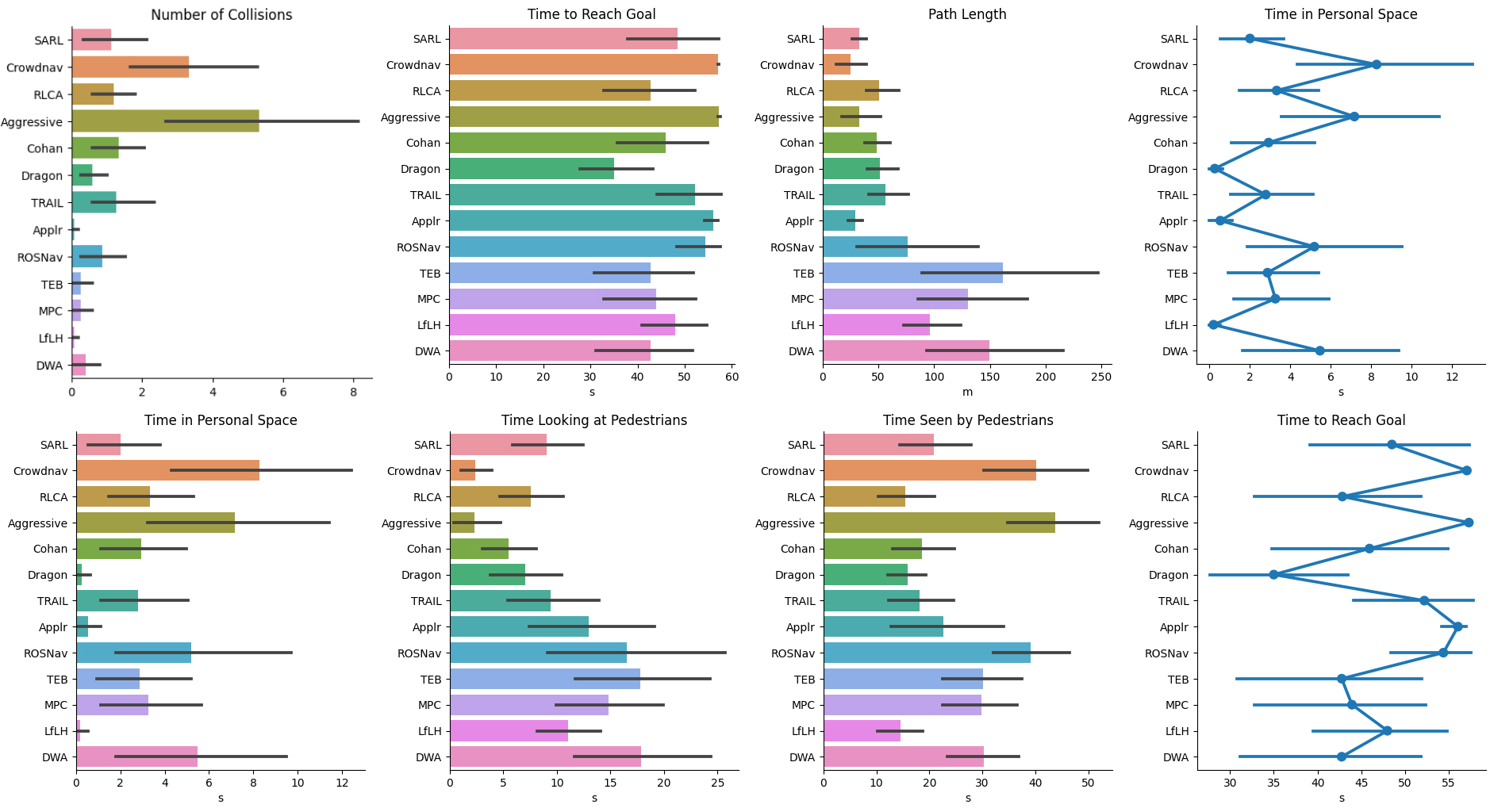}
    \caption{Sample plots for the benchmark runs conducted using the benchmark tasks. Thereby, the average values over all episodes are calculated and can be plotted in different representations}
    \label{fig:plots}
\end{figure*}

\subsection{User Study}
\noindent
We evaluated our platform by conducting a user study involving 30 participants from universities in Germany, the US, Singapore, Japan, and Korea with varying levels of expertise in robotics. 
To ensure a wide range of variety, the participants came from one of the following groups: students and researchers who have worked with one of the previous versions of Arena; researchers in social navigation that we know from international conferences who have never worked with Arena before; researchers from our lab who have never worked with Arena before; computer science, computer engineering, and electrical engineering students from different international universities who we asked to participate who have never worked with Arena before.
The participants were asked to install and interact with the platform by executing a list of tasks ranging from starting different task modes, starting a training, and running different planners on different robots. Subsequently, they were asked to answer a questionnaire, which included questions such as describing the pros and cons of using the platform, general thoughts, improvement ideas, issues and complexity of operation. Furthermore, the participants also were asked weather they have worked with Arena before and what their occupation and experience in robotics is.
All responses and information about the participants occupation and level of expertise are publicly available on \href{https://drive.google.com/drive/folders/1lPJrvgxsfZyrIH3be6EFNX4R7y3PkL3J?usp=sharing}{\color{blue} Google Drive}. 
\\\noindent
The feedback predominantly highlighted the positive developments in Arena 3.0 compared to its predecessors. Notably, users appreciated the streamlined installation process, citing it as a significant improvement that enhances the platform's accessibility. A substantial enhancement was observed in the task generator's robustness, which participants found to enhance the platform's overall utility significantly compared to previous versions. Moreover, the inclusion of a diverse array of robots and planners was well-received, with users expressing satisfaction with the breadth of options now available for experimentation. This variety was seen as a crucial factor in facilitating a wider range of research and development activities. Regarding the future utilization of Arena 3.0, an overwhelming majority of the respondents expressed a keen interest in continuing to use the platform. When asked about specific applications, participants mentioned areas such as autonomous navigation in dynamic environments, multi-robot coordination, and testing of novel navigation algorithms as primary use-cases.
\\\noindent
Despite the positive reception, users also offered constructive criticism, pointing out areas for further improvement. The primary downsides identified included a perceived learning curve for new users, particularly in navigating advanced features such as using the API to integrate new task modes/planners or development using the Unity simulator. Some users pointed out the need for more documentation to aid in this process. Additionally, users stated that the requirement to use Ubuntu 20.04 is a major limitation. 
Further, it is to note that 14 out of the 30 participants had previously worked with one of the previous versions of Arena, which could possibly influence their assessment positively as they may be excited about the new functions introduced. However, we also observed that users who were familiar with the previous versions reported that they were impacted negatively by extensive modifications of the code base and API.
\\\noindent
Nevertheless, the user study substantiated the value of the improvements made in Arena 3.0. The platform is more robust, versatile, and user-friendly, which can significantly contribute to robotic navigation research. The highlighted remarks and issues resulting from the user study will be incorporated in the next iteration of Arena. This includes an ongoing integration and migration of ROS2, a more profound documentation, optimized installation processes, as well as improving the companion web-application Arena-web \cite{kastner2023arenaweb} to feature a majority of functionalities with the objective to make installation of Arena obsolete in future.

\subsection{Navigation Benchmark}
\noindent
To showcase the benchmark capabilities of Arena 3.0, we performed a series of evaluation runs using a Jackal robot in the newly-introduced benchmark task mode. The configuration for these evaluations was structured to cover a wide range of scenarios. It included 100 episodes of randomly generated environments with static obstacles and up to 30 pedestrians, followed by 10 episodes in each of the specially designed environments - canteen, industrial hall, office (both indoor and outdoor maps), and a targeted corridor scenario. These evaluation runs are shown in the supplementary video accompanying this paper.
\\\noindent
The metrics chosen for this demonstration were path length and time to goal, which serve as indicators of the planner's efficiency. Collision rates were monitored as a measure of safety. Alongside these standard metrics, we also incorporated the three social metrics to evaluate the robot's interaction with pedestrians: time spent in private space, time visible to pedestrians, and time spent facing pedestrians. These social metrics provide a deeper insight into the robot's performance in socially complex environments and interactions.
\noindent
Exemplary plots depicting the evaluation results are presented in Figure \ref{fig:plots}. These plots highlight the performance of different planners based on social metrics and efficiency. For example, the LfLH planner, which is specifically tailored for social environments, exhibits better performance in terms of the social metrics. However, it demonstrates lower efficiency compared to other planners. In contrast, our Rosnav planner, while efficient, tends to infringe on pedestrian privacy for longer durations, as indicated by an average time of 5.6 s. This is attributed to the fact that Rosnav was primarily developed with a focus on efficiency, without the integration of social metrics. These findings underscore the potential for further refinement of the Rosnav planner, particularly in its reward system, to enhance its social navigation capabilities. Overall, the evaluations conducted using Arena 3.0 showcase its effectiveness for extensive benchmarking, demonstrating its capacity to produce reliable results across high numbers of episodes. This robustness underscores the platform's suitability for detailed and comprehensive analysis of planner performance in various scenarios.

%% file: 5-conclusion.tex
\section{Conclusion}
\noindent 
In this paper, we introduced Arena 3.0, an enhanced version of our previous platforms, Arena-Bench and Arena 1.0, and Arena 2.0. The platform integrates realistic pedestrian movements and social force modeling, including human-human and human-robot interactions. Other significant features are the dynamic world generation of common social areas such as canteens, offices, or industry halls, an improved task generator with a wider array of customizable and dynamically adjustable task modes, and an expanded set of robots and state-of-the-art planners. We included new evaluation metrics focused on social navigation aspects, enhanced APIs for ease of integration, and state-of-the-art social force modeling for realistic crowd simulations. The platform's navigation framework has been refined, and the abstraction of simulators extends the range of potential users and possibilities. 
\\\noindent
Future works include the migration towards ROS2, integration of drones, or the employment of multi-agent-reinforcement-learning approaches. Further, we aspire to release our second version of Arena-web \cite{kastner2023arenaweb}, which will offer a majority of the current platforms functionalities within a web-application making installation of the platform obsolete and facilitating development and testing for a wider audience. Finally, the extension of human-object interactions such as humans interacting with furniture or other objects to create more realistic situations, will be integrated in future works.

%% file: acknowledgment.tex

\section*{Acknowledgements}

\noindent This research / project is supported by A*STAR under its National Robotics Programme (NRP) (Award M23NBK0053). This research is supported in part by the National Research Foundation (NRF), Singapore and DSO National Laboratories under the AI Singapore Program (Award Number: AISG2-RP-2020-017).

%% file: main.bbl
\begin{thebibliography}{10}
\providecommand{\url}[1]{#1}
\csname url@samestyle\endcsname
\providecommand{\newblock}{\relax}
\providecommand{\bibinfo}[2]{#2}
\providecommand{\BIBentrySTDinterwordspacing}{\spaceskip=0pt\relax}
\providecommand{\BIBentryALTinterwordstretchfactor}{4}
\providecommand{\BIBentryALTinterwordspacing}{\spaceskip=\fontdimen2\font plus
\BIBentryALTinterwordstretchfactor\fontdimen3\font minus \fontdimen4\font\relax}
\providecommand{\BIBforeignlanguage}[2]{{%
\expandafter\ifx\csname l@#1\endcsname\relax
\typeout{** WARNING: IEEEtran.bst: No hyphenation pattern has been}%
\typeout{** loaded for the language `#1'. Using the pattern for}%
\typeout{** the default language instead.}%
\else
\language=\csname l@#1\endcsname
\fi
#2}}
\providecommand{\BIBdecl}{\relax}
\BIBdecl

\bibitem{kastner2022arena-bench}
L.~K{\"a}stner, T.~Bhuiyan, T.~A. Le, E.~Treis, J.~Cox, B.~Meinardus, J.~Kmiecik, R.~Carstens, D.~Pichel, B.~Fatloun \emph{et~al.}, ``Arena-bench: A benchmarking suite for obstacle avoidance approaches in highly dynamic environments,'' \emph{IEEE Robotics and Automation Letters}, vol.~7, no.~4, pp. 9477--9484, 2022.

\bibitem{kastner2021arena}
L.~K{\"a}stner, T.~Buiyan, L.~Jiao, T.~A. Le, X.~Zhao, Z.~Shen, and J.~Lambrecht, ``Arena-rosnav: Towards deployment of deep-reinforcement-learning-based obstacle avoidance into conventional autonomous navigation systems,'' in \emph{2021 IEEE/RSJ International Conference on Intelligent Robots and Systems (IROS)}.\hskip 1em plus 0.5em minus 0.4em\relax IEEE, 2021, pp. 6456--6463.

\bibitem{kastner2023arena}
L.~K{\"a}stner, R.~Carstens, H.~Zeng, J.~Kmiecik, T.~Bhuiyan, N.~Khorsandhi, V.~Shcherbyna, and J.~Lambrecht, ``Arena-rosnav 2.0: A development and benchmarking platform for robot navigation in highly dynamic environments,'' in \emph{2023 IEEE/RSJ International Conference on Intelligent Robots and Systems (IROS)}.\hskip 1em plus 0.5em minus 0.4em\relax IEEE, 2023, pp. 11\,257--11\,264.

\bibitem{tsoi2022sean}
N.~Tsoi, A.~Xiang, P.~Yu, S.~S. Sohn, G.~Schwartz, S.~Ramesh, M.~Hussein, A.~W. Gupta, M.~Kapadia, and M.~V{\'a}zquez, ``Sean 2.0: Formalizing and generating social situations for robot navigation,'' \emph{IEEE Robotics and Automation Letters}, vol.~7, no.~4, pp. 11\,047--11\,054, 2022.

\bibitem{francis2023principles}
A.~Francis, C.~P{\'e}rez-d'Arpino, C.~Li, F.~Xia, A.~Alahi, R.~Alami, A.~Bera, A.~Biswas, J.~Biswas, R.~Chandra \emph{et~al.}, ``Principles and guidelines for evaluating social robot navigation algorithms,'' \emph{arXiv preprint arXiv:2306.16740}, 2023.

\bibitem{everett2018motion}
M.~Everett, Y.~F. Chen, and J.~P. How, ``Motion planning among dynamic, decision-making agents with deep reinforcement learning,'' in \emph{2018 IEEE/RSJ International Conference on Intelligent Robots and Systems (IROS)}.\hskip 1em plus 0.5em minus 0.4em\relax IEEE, 2018, pp. 3052--3059.

\bibitem{move_base_flex}
``{Move Base Flex},'' \url{https://github.com/magazino/move_base_flex}, accessed: 2024-01-22.

\bibitem{dugas2020navrep}
D.~Dugas, J.~Nieto, R.~Siegwart, and J.~J. Chung, ``Navrep: Unsupervised representations for reinforcement learning of robot navigation in dynamic human environments,'' in \emph{2021 IEEE International Conference on Robotics and Automation (ICRA)}, 2021, pp. 7829--7835.

\bibitem{long2017deep}
P.~Long, W.~Liu, and J.~Pan, ``Deep-learned collision avoidance policy for distributed multiagent navigation,'' \emph{IEEE Robotics and Automation Letters}, vol.~2, no.~2, pp. 656--663, 2017.

\bibitem{xiao2017human}
Q.~Xiao, F.~Sun, R.~Ge, K.~Chen, and B.~Wang, ``Human tracking and following of mobile robot with a laser scanner,'' in \emph{2017 2nd International Conference on Advanced Robotics and Mechatronics (ICARM)}.\hskip 1em plus 0.5em minus 0.4em\relax IEEE, 2017, pp. 675--680.

\bibitem{rosmann2015planning}
C.~R{\"o}smann, F.~Hoffmann, and T.~Bertram, ``Planning of multiple robot trajectories in distinctive topologies,'' in \emph{2015 European Conference on Mobile Robots (ECMR)}.\hskip 1em plus 0.5em minus 0.4em\relax IEEE, 2015, pp. 1--6.

\bibitem{singamaneni2021human}
P.~T. Singamaneni, A.~Favier, and R.~Alami, ``Human-aware navigation planner for diverse human-robot ineraction contexts,'' in \emph{IEEE/RSJ International Conference on Intelligent Robots and Systems (IROS)}, 2021.

\bibitem{tsoi2020sean}
N.~Tsoi, M.~Hussein, J.~Espinoza, X.~Ruiz, and M.~V{\'a}zquez, ``Sean: Social environment for autonomous navigation,'' in \emph{Proceedings of the 8th International Conference on Human-Agent Interaction}, 2020, pp. 281--283.

\bibitem{chen2019crowd}
C.~Chen, Y.~Liu, S.~Kreiss, and A.~Alahi, ``Crowd-robot interaction: Crowd-aware robot navigation with attention-based deep reinforcement learning,'' in \emph{2019 International Conference on Robotics and Automation (ICRA)}.\hskip 1em plus 0.5em minus 0.4em\relax IEEE, 2019, pp. 6015--6022.

\bibitem{chen2017socially}
Y.~F. Chen, M.~Everett, M.~Liu, and J.~P. How, ``Socially aware motion planning with deep reinforcement learning,'' in \emph{2017 IEEE/RSJ International Conference on Intelligent Robots and Systems (IROS)}.\hskip 1em plus 0.5em minus 0.4em\relax IEEE, 2017, pp. 1343--1350.

\bibitem{kastner-aio}
L.~Kästner, J.~Cox, T.~Buiyan, and J.~Lambrecht, ``All-in-one: A drl-based control switch combining state-of-the-art navigation planners,'' in \emph{2022 International Conference on Robotics and Automation (ICRA)}, 2022, pp. 2861--2867.

\bibitem{bench_mr}
E.~Heiden, L.~Palmieri, L.~Bruns, K.~O. Arras, G.~S. Sukhatme, and S.~Koenig, ``Bench-mr: A motion planning benchmark for wheeled mobile robots,'' \emph{IEEE Robotics and Automation Letters}, vol.~6, no.~3, pp. 4536--4543, 2021.

\bibitem{robobench}
J.~Weisz, Y.~Huang, F.~Lier, S.~Sethumadhavan, and P.~Allen, ``Robobench: Towards sustainable robotics system benchmarking,'' in \emph{2016 IEEE International Conference on Robotics and Automation (ICRA)}, 2016, pp. 3383--3389.

\bibitem{althoff2017commonroad}
M.~Althoff, M.~Koschi, and S.~Manzinger, ``Commonroad: Composable benchmarks for motion planning on roads,'' in \emph{2017 IEEE Intelligent Vehicles Symposium (IV)}.\hskip 1em plus 0.5em minus 0.4em\relax IEEE, 2017, pp. 719--726.

\bibitem{moll2015benchmarking}
M.~Moll, I.~A. Sucan, and L.~E. Kavraki, ``Benchmarking motion planning algorithms: An extensible infrastructure for analysis and visualization,'' \emph{IEEE Robotics \& Automation Magazine}, vol.~22, no.~3, pp. 96--102, 2015.

\bibitem{holtz2021socialgym}
J.~Holtz and J.~Biswas, ``Socialgym: A framework for benchmarking social robot navigation,'' \emph{arXiv preprint arXiv:2109.11011}, 2021.

\bibitem{perezhunavsim}
N.~P{\'e}rez-Higueras, R.~Otero, F.~Caballero, and L.~Merino, ``Hunavsim: A ros2 human navigation simulator for benchmarking human-aware robot navigation,'' \emph{cit. on}, p.~49.

\bibitem{mrpb}
J.~Wen, X.~Zhang, Q.~Bi, Z.~Pan, Y.~Feng, J.~Yuan, and Y.~Fang, ``Mrpb 1.0: A unified benchmark for the evaluation of mobile robot local planning approaches,'' in \emph{2021 IEEE International Conference on Robotics and Automation (ICRA)}, 2021, pp. 8238--8244.

\bibitem{xia2020interactive}
F.~Xia, W.~B. Shen, C.~Li, P.~Kasimbeg, M.~E. Tchapmi, A.~Toshev, R.~Mart{\'\i}n-Mart{\'\i}n, and S.~Savarese, ``Interactive gibson benchmark: A benchmark for interactive navigation in cluttered environments,'' \emph{IEEE Robotics and Automation Letters}, vol.~5, no.~2, pp. 713--720, 2020.

\bibitem{portugal2019ros}
D.~Portugal, L.~Iocchi, and A.~Farinelli, ``A ros-based framework for simulation and benchmarking of multi-robot patrolling algorithms,'' in \emph{Robot Operating System (ROS)}.\hskip 1em plus 0.5em minus 0.4em\relax Springer, 2019, pp. 3--28.

\bibitem{gao2022evaluation}
Y.~Gao and C.-M. Huang, ``Evaluation of socially-aware robot navigation,'' \emph{Frontiers in Robotics and AI}, vol.~8, p. 420, 2022.

\bibitem{xiao2022autonomous}
X.~Xiao, Z.~Xu, Z.~Wang, Y.~Song, G.~Warnell, P.~Stone, T.~Zhang, S.~Ravi, G.~Wang, H.~Karnan \emph{et~al.}, ``Autonomous ground navigation in highly constrained spaces: Lessons learned from the benchmark autonomous robot navigation challenge at icra 2022 [competitions],'' \emph{IEEE Robotics \& Automation Magazine}, vol.~29, no.~4, pp. 148--156, 2022.

\bibitem{nair2022dynabarn}
A.~Nair, F.~Jiang, K.~Hou, Z.~Xu, S.~Li, X.~Xiao, and P.~Stone, ``Dynabarn: Benchmarking metric ground navigation in dynamic environments,'' in \emph{2022 IEEE International Symposium on Safety, Security, and Rescue Robotics (SSRR)}.\hskip 1em plus 0.5em minus 0.4em\relax IEEE, 2022, pp. 347--352.

\bibitem{redmon2016you}
J.~Redmon, S.~Divvala, R.~Girshick, and A.~Farhadi, ``You only look once: Unified, real-time object detection,'' in \emph{Proceedings of the IEEE conference on computer vision and pattern recognition}, 2016, pp. 779--788.

\bibitem{kastner2023arenaweb}
L.~K{\"a}stner, R.~Carstens, L.~Nahrwold, C.~Liebig, V.~Shcherbyna, S.~Lee, and J.~Lambrecht, ``Demonstrating arena-web: A web-based development and benchmarking platform for autonomous navigation approaches,'' \emph{Robotics: Science and Systems (RSS)}, 2023.

\bibitem{helbing1995social}
D.~Helbing and P.~Molnar, ``Social force model for pedestrian dynamics,'' \emph{Physical review E}, vol.~51, no.~5, p. 4282, 1995.

\bibitem{pysocialforce}
``{PySocialForce},'' \url{https://github.com/yuxiang-gao/PySocialForce}, accessed: 2024-01-22.

\bibitem{deepsocialforce}
S.~Kreiss, ``Deep social force,'' 2021.

\bibitem{sfm-evacuation}
``{Evacuation-Bottleneck},'' \url{https://github.com/fschur/Evacuation-Bottleneck}, accessed: 2024-01-22.

\bibitem{sfm-bonding}
\BIBentryALTinterwordspacing
S.~Xu and H.~B.-L. Duh, ``A simulation of bonding effects and their impacts on pedestrian dynamics,'' \emph{Trans. Intell. Transport. Syst.}, vol.~11, no.~1, p. 153–161, mar 2010. [Online]. Available: \url{https://doi.org/10.1109/TITS.2009.2036152}
\BIBentrySTDinterwordspacing

\bibitem{van2011reciprocal}
J.~Van Den~Berg, J.~Snape, S.~J. Guy, and D.~Manocha, ``Reciprocal collision avoidance with acceleration-velocity obstacles,'' in \emph{2011 IEEE International Conference on Robotics and Automation}.\hskip 1em plus 0.5em minus 0.4em\relax IEEE, 2011, pp. 3475--3482.

\bibitem{van2011rvo2}
J.~Van Den~Berg, S.~J. Guy, M.~Lin, and D.~Manocha, ``Reciprocal collision avoidance with acceleration-velocity obstacles,'' \emph{2011 IEEE International Conference on Robotics and Automation}, pp. 3475--3482, 2011.

\bibitem{sfm-moussaid}
\BIBentryALTinterwordspacing
M.~Moussaïd, N.~Perozo, S.~Garnier, D.~Helbing, and G.~Theraulaz, ``The walking behaviour of pedestrian social groups and its impact on crowd dynamics,'' \emph{PLOS ONE}, vol.~5, no.~4, pp. 1--7, 04 2010. [Online]. Available: \url{https://doi.org/10.1371/journal.pone.0010047}
\BIBentrySTDinterwordspacing

\bibitem{sfm-helbing}
\BIBentryALTinterwordspacing
D.~Helbing and P.~Moln\'ar, ``Social force model for pedestrian dynamics,'' \emph{Phys. Rev. E}, vol.~51, pp. 4282--4286, May 1995. [Online]. Available: \url{https://link.aps.org/doi/10.1103/PhysRevE.51.4282}
\BIBentrySTDinterwordspacing

\bibitem{xiao2020appld}
X.~Xiao, B.~Liu, G.~Warnell, J.~Fink, and P.~Stone, ``Appld: Adaptive planner parameter learning from demonstration,'' \emph{IEEE Robotics and Automation Letters}, vol.~5, no.~3, pp. 4541--4547, 2020.

\bibitem{khatib1986real}
O.~Khatib, ``Real-time obstacle avoidance for manipulators and mobile robots,'' in \emph{Autonomous robot vehicles}.\hskip 1em plus 0.5em minus 0.4em\relax Springer, 1986, pp. 396--404.

\bibitem{rosmann2015timed}
C.~R{\"o}smann, F.~Hoffmann, and T.~Bertram, ``Timed-elastic-bands for time-optimal point-to-point nonlinear model predictive control,'' in \emph{2015 european control conference (ECC)}.\hskip 1em plus 0.5em minus 0.4em\relax IEEE, 2015, pp. 3352--3357.

\bibitem{rosmann2019time}
C.~R{\"o}smann, ``Time-optimal nonlinear model predictive control,'' Ph.D. dissertation, Dissertation, Technische Universit{\"a}t Dortmund, 2019.

\bibitem{xiao2022motion}
X.~Xiao, B.~Liu, G.~Warnell, and P.~Stone, ``Motion planning and control for mobile robot navigation using machine learning: a survey,'' \emph{Autonomous Robots}, pp. 1--29, 2022.

\bibitem{long2018towards}
P.~Long, T.~Fan, X.~Liao, W.~Liu, H.~Zhang, and J.~Pan, ``Towards optimally decentralized multi-robot collision avoidance via deep reinforcement learning,'' in \emph{2018 IEEE International Conference on Robotics and Automation (ICRA)}.\hskip 1em plus 0.5em minus 0.4em\relax IEEE, 2018, pp. 6252--6259.

\bibitem{li2019sarl}
K.~Li, Y.~Xu, J.~Wang, and M.~Q.-H. Meng, ``Sarl-star: Deep reinforcement learning based human-aware navigation for mobile robot in indoor environments,'' in \emph{2019 IEEE International Conference on Robotics and Biomimetics (ROBIO)}.\hskip 1em plus 0.5em minus 0.4em\relax IEEE, 2019, pp. 688--694.

\bibitem{muchen23}
M.~Sun, F.~Baldini, P.~Trautman, and T.~Murphey, ``Move beyond trajectories: Distribution space coupling for crowd navigation,'' in \emph{Robotics: Science and Systems (RSS)}, 2021.

\bibitem{yangiplanner23}
F.~Yang, C.~Wang, C.~Cadena, and M.~Hutter, ``iplanner: Imperative path planning,'' in \emph{Robotics: Science and Systems (RSS)}, 2023.

\bibitem{wellhausen2021rough}
L.~Wellhausen and M.~Hutter, ``Rough terrain navigation for legged robots using reachability planning and template learning,'' in \emph{2021 IEEE/RSJ International Conference on Intelligent Robots and Systems (IROS 2021)}, 2021.

\bibitem{move_base}
``{Move Base},'' \url{http://wiki.ros.org/move_base}, accessed: 2024-01-22.

\bibitem{ros2}
\BIBentryALTinterwordspacing
S.~Macenski, T.~Foote, B.~Gerkey, C.~Lalancette, and W.~Woodall, ``Robot operating system 2: Design, architecture, and uses in the wild,'' \emph{Science Robotics}, vol.~7, no.~66, p. eabm6074, 2022. [Online]. Available: \url{https://www.science.org/doi/abs/10.1126/scirobotics.abm6074}
\BIBentrySTDinterwordspacing

\bibitem{kastner2021connecting}
L.~K{\"a}stner, X.~Zhao, T.~Buiyan, J.~Li, Z.~Shen, J.~Lambrecht, and C.~Marx, ``Connecting deep-reinforcement-learning-based obstacle avoidance with conventional global planners using waypoint generators,'' in \emph{2021 IEEE/RSJ International Conference on Intelligent Robots and Systems (IROS)}.\hskip 1em plus 0.5em minus 0.4em\relax IEEE, pp. 1213--1220.

\bibitem{rubagotti2022perceived}
M.~Rubagotti, I.~Tusseyeva, S.~Baltabayeva, D.~Summers, and A.~Sandygulova, ``Perceived safety in physical human--robot interaction—a survey,'' \emph{Robotics and Autonomous Systems}, vol. 151, p. 104047, 2022.

\bibitem{howell2023effects}
P.~Howell, J.~Kolb, Y.~Liu, and H.~Ravichandar, ``The effects of robot motion on comfort dynamics of novice users in close-proximity human-robot interaction,'' in \emph{2023 IEEE/RSJ International Conference on Intelligent Robots and Systems (IROS)}.\hskip 1em plus 0.5em minus 0.4em\relax IEEE, 2023, pp. 9859--9864.

\bibitem{wang2019learning}
W.~Wang, Y.~Chen, R.~Li, and Y.~Jia, ``Learning and comfort in human--robot interaction: A review,'' \emph{Applied Sciences}, vol.~9, no.~23, p. 5152, 2019.

\end{thebibliography}
